# Wavelet-Based Iterative Learning Control with Fuzzy PD Feedback for Position Tracking of A Pneumatic Servo System

C. E. Huang* and J. S. Chen⁼

*Department of Power Mechanical Engineering
National Tsing-Hua University
Hsinchu, Taiwan.
e-mail: d947710@oz.nthu.edu.tw

⁼Department of Power Mechanical Engineering
National Tsing-Hua University
Hsinchu, Taiwan.
e-mail: jschen@pme.nthu.edu.tw

**Abstract**

In this paper, a wavelet-based iterative learning control (WILC) scheme with Fuzzy PD feedback is presented for a pneumatic control system with nonsmooth nonlinearities and uncertain parameters. The wavelet transform is employed to extract the learnable dynamics from measured output signal before it can be used to update the control profile. The wavelet transform is adopted to decompose the original signal into many low-resolution signals that contain the learnable and unlearnable parts. The desired control profile is then compared with the learnable part of the transformed signal. Thus, the effects from unlearnable dynamics on the controlled system can be attenuated by a Fuzzy PD feedback controller. As for the rules of Fuzzy PD controller in the feedback loop, a genetic algorithm (GA) is employed to search for the inference rules of optimization. A proportional-valve controlled pneumatic cylinder actuator system is used as the control target for simulation. Simulation results have shown a much-improved position-tracking performance.

## 1 Introduction

Most physical components posses nonsmooth nonlinearities, such as saturation, asymmetric dynamic, friction, and deadzone. Such nonlinearities are usually seen in actuators used in practice, such as pneumatic valve. Pneumatic actuation systems are usually used in industrial machines and automation applications because of their enthralling advantages, e.g., clean environment, easy to use, high performance due to low power-weight ratio, leanness, low price, etc. Moreover, certain disadvantages of pneumatic-actuated systems such as compressibility of the working fluid, dead zone and stick friction can be overcome by certain control strategies, either. The goal of this work is thus to propose an effective position-tracking controller for a proportional-valve controlled pneumatic positioning system with repeated trajectory.

Iterative learning control (ILC), first introduced by Arimoto *et al* [1], has been proved to be a very efficient control methodology, because it can improve tracking control performance through repeated trials. The ILC scheme as reported by [1] is a feedforward action in nature completely based on the previous cycle error of a repeated task, and the resulting control system is basically an open-loop system. It can also handle nonlinear systems as well as improve the tracking accuracy, its applications such as robot trajectory control, position control of mechanical systems [2] and pneumatic actuated system [3] etc., usually involve tasks of repetitive behavior. Amann *et al* [4], Bien[5], Kurek [6] and Moore *et al* [7] also reported he development of different learning control schemes. Although theoretical proof for the applicability of the feedforward only learning control scheme, which is also called previous cycle error (PCE) type ILC scheme [5], has been reported, it still suffers from the following shortcomings. If the open-loop system is unstable and is not robust against disturbances that are not repeatable among twelve iterations, adverse effects might be exhibited from the output of ILC. These feedback-based ILC schemes are also named as ILC of current cycle error (CCE) type. The convergence of tracking error between the system output and a reference input is ensured by the proposition of a PID-type learning algorithm. On the other hand, wavelet scheme was first successfully applied in ILC by Tseng and Chen [8] in flexible mechanism with a simple proportional feedback loop. Later, inspired by [8], Chien *et al* [9] also applied wavelet transform filtering scheme on an enhanced ILC for the trajectory tracking of a piezoelectric-driven system, the wavelet transform is served as a filter to achieve tracking errors with zero phase. The experimental results have shown that within several iterations, the tracking accuracies of the proposed scheme would have the steady-state errors that are very close to the system noise level. Hsu *et al* [10] also proposed a neural backstepping control scheme where the controller consisted a wavelet neural network (WNN) identifier. This controller can handle systems with parameter variations and unknown dynamics. Furthermore, a $L_2$ robust control is designed to achieve tracking performance with desired attenuation level. This kind of hybrid controller can achieve favorable tracking performance of a chaotic system and a wing-rock motion system.

Although many nonlinear systems can be handled by the feedback-based ILC controller that has been provided satisfactory performances, the design of a feedback-based ILC controller still encountered some obstacles. The major obstacles are that the system must be learnable, the system dynamics must be time-invariant or slowly time-varying and the external disturbances are repeatable during iterative process. However, in most practical system, there exhibit a lot of non-repeatable disturbances and noises that may contaminate the system input and the measured signal respectively. Nevertheless, the uncertain dynamics due to the non-smooth nonlinearities would cause a large tracking error. While using ILC in controlling a nonlinear dynamic system, the unexpected vibration may also be excited by the nonlinear disturbances and nonlinearities. The amplitude of vibration will rise rapidly during the process of iterative process. Namely, it would corrupt the control profile and cause instability during the iterative operation. The main objective of this paper is to extend the results in the author's previous work [8] to develop an efficient ILC scheme that incorporated with a much sophisticated feedback control the enhance the performance of a pneumatic servo system with unlearnable dynamics

## 2 Dynamic system model of pneumatic cylinder

A typical single-rod pneumatic servo system with a proportional valve is employed as the control target under study verification. The configuration of the proposed controlled system is as shown in Figure 1. Saying adiabatic charging and discharging of the cylinder chambers [11], the dynamic equations of nonlinear Pneumatic actuation system can be describe as [12]

$$\begin{aligned}
\dot{x}_p &= v_p \\
\dot{v}_p &= \frac{1}{M}\left(-bv_p + A_a P_a - A_b P_b - F_f - F_L\right) \\
\dot{P}_a &= \frac{1}{V_a}\left(\gamma R T \dot{m}_a - \beta\gamma P_a A_a v_p\right) \\
\dot{P}_b &= \frac{1}{V_b}\left(-\gamma R T \dot{m}_b + \beta\gamma P_b A_b v_p\right) \\
\dot{x}_v &= -\frac{1}{\tau}\left(x_v - k_v u\right)
\end{aligned} \qquad (1)$$

Where, $x_p$ is the position of the cylinder and $v_p$ is the velocity of the cylinder. $\dot{m}_a$, $\dot{m}_b$, $A_a$, $A_b$, $P_a$, $P_b$, $V_a$ and $V_b$ are the mass flow rate of air through each control valve orifice, piston annulus areas, the instantaneous cylinder chamber absolute pressures and volumes, respectively. $M$ is total mass of piston, rods, and load. $R$ is ideal gas constant. $T$ is temperature of air source. Parameter $k_v$ is the valve spool position gain. Parameter $\tau$ is the valve first-order time constant. Parameter $\gamma$ is ratio of specific heats. Parameter $b$ is viscous damping coefficient. Parameter $\beta$ is a compressibility flow correction factor, which accounts for the fact that the pressure-volume work process is neither adiabatic nor isothermal but somewhere in between [13]. $F_f$ represents the dry friction force and $F_L$ signifies the externally applied load.

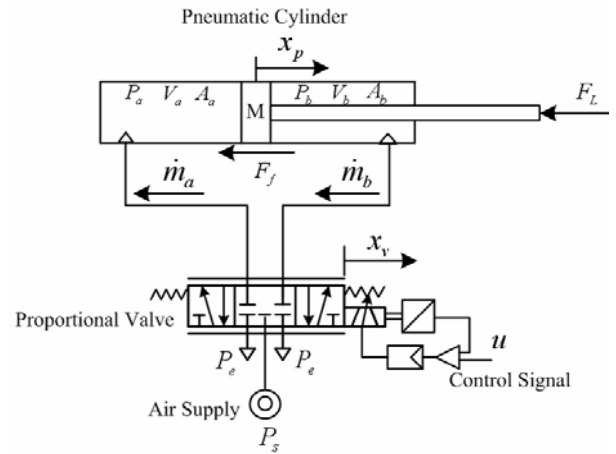

**Figure 1: The configuration of the valve-controlled pneumatic system**

Additionally, the manufacturer suggests that the dynamics of the control valve spool is modeled as a first-order system where the displacement of the valve spool is denoted by $x_v$ and $u$ is the control signal. The mass flow rate of air through each control valve orifice is handled by the nonlinear dynamic equation [14]:

$$\dot{m} = \begin{cases} \dfrac{C_1 C_d w x_v P_u}{\sqrt{T}} & \text{if } \dfrac{P_d}{P_u} \le P_{cr} \\[2ex] \dfrac{C_1 C_d w x_v P_u}{\sqrt{T}}\sqrt{1-\left(\dfrac{P_d/P_u - P_{cr}}{1-P_{cr}}\right)^2} & \text{if } \dfrac{P_d}{P_u} > P_{cr} \end{cases} \qquad (2)$$

Where $C_1 = \sqrt{\dfrac{\gamma}{R}\left(\dfrac{2}{\gamma+1}\right)^{(\gamma+1)/(\gamma-1)}}$

In equation (2), $w$ is valve orifice area gradient, $P_d$ is the absolute downstream pressure, while $P_u$ denotes the absolute upstream pressure and $P_{cr}$ is critical pressure ratio. From the dynamic equations given above, we known that a pneumatic system is essential unsymmetrical system because of different effective piston areas in the lift- and right-hand sides of the cylinder. However, the linear controller with constant-gain such as PID can not track the desired position reference accurately. Therefore, a WILC with Fuzzy PD feedback is used to control the pneumatic system to follow the desired trajectory.

### 3 The proposed controller design

The learning law can update the control profile through calculating the previous learning control profile, previous cycle error and current cycle error, but the system must satisfy the property of repeatable or learnable. In other words, the system disturbance must be repeatable during the iterative process and the system dynamics must possess invariant dynamics. These properties are difficult to satisfy in pneumatic actuation systems. To enhance the tracking performance, a wavelet-transform process is utilized in suppressing the unlearnable system dynamics where the wavelet transform could decompose the original signal into learnable and unlearnable parts. The learnable parts are used as the feedforward information in repetitive operation, while the desired control profile is then compared with the learnable part of the transformed signal, and improves the tracking performance in progressive fashion. On the other hand, the effects from unlearnable dynamics on the controlled system can be attenuated by a Fuzzy PD feedback controller where the inference rules are optimized by GA algorithm.

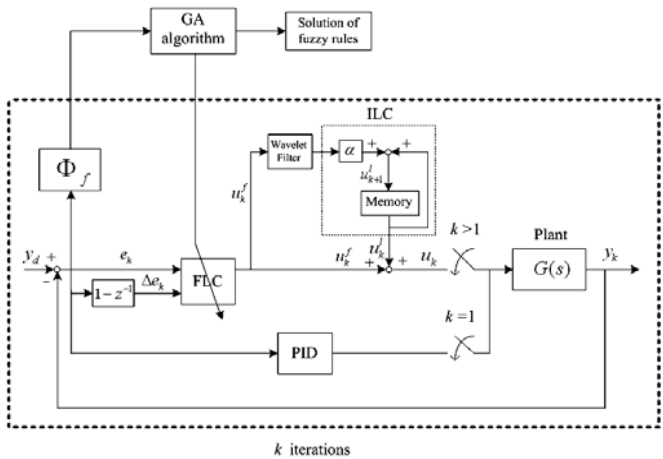

**Figure 2 The proposed controller with GA**

It is noted that a PID controller is augmented to control the system when $k=1$, i.e. the first iteration, to ensure the same initial root-mean-square (RMS) value in the beginning of every fitness function calculation of GA algorithm. While the control parameters of this PID controller are obtained based on the stability limit criterion in [15]. The criterion used a proportional controller with gain $K_u$ to control the system with a step reference input, and then increasing the proportional gain until a sustained oscillations are observed,. Then, with the ultimate gain $K_u$ and the period of oscillation $T_u$ the control parameters can be determined as shown below.

$$K_P = 0.6K_u, \quad K_I = \frac{1.2K_u}{T_u} \quad \text{and} \quad K_D = 0.075K_uT_u \quad (3)$$

### 3.1 The feedforward loop - a WILC design

In this subsection, a WILC design will be described, its block-diagram of the controlled system is as shown in Figure 4, where $y_d$ is the desired output trajectory and $y_k$ is the system output of the $k$ th iteration. The control scheme is expressed as follows:

$$u_k(t) = u_k^l(t) + u_k^f(t) \quad (4)$$

$$u_k^l(t) = u_{k-1}^l(t) + \alpha W * u_{k-1}^f(t) \quad (5)$$

$$u_k^f(t) = f(e_k(t), \Delta e_k(t)) \quad (6)$$

Where $\alpha$ are positive and fixed learning gain, the $u_k^f$ is a Fuzzy PD feedback controller output for the $k$ th iteration, $u_k^l$ is an WILC output for the $k$ th iteration and $W*$ stands for the wavelet transform as a filtering operator in time domain. The feedback control signal $u_k^f$ is filtered by wavelet transform and then utilized to update the learning profile.

There are many applications using Wavelets successfully in a wide variety of research areas such as image processing, signal analysis and de-noising [16]. In this study the multi-resolution analysis (MRA) is utilized to extract the unlearnable dynamics apart from an ILC system and then apply the resultant ILC control effort with the feedback control effort that attempts to attenuate the unlearnable dynamics. The MRA is a tool that uses the wavelet transform to map a one-dimensional time signal into a two-dimensional signal represented in both the frequency and time domain. A time function $f(t)$ transferred by the wavelet with a function $\psi(t)$, called mother wavelet, is defined as the inner product of $f(t)$ with $\psi_{a,b}(t)$, i.e.

$$Wf(a,b) = \langle f(t), \psi_{a,b} \rangle \quad (7)$$

Where $a$ and $b$ are a scaling factor and a shift parameter, respectively.

The signal $f(t)$ can then be decomposed into several lower-resolution components. A typical wavelet decomposition tree for a time-domain signal $f(t)$ is shown in Figure 3, where $A_n$ and $D_j, j = 1, 2, ..., n$ are denoted as the approximations and details, standing for low-frequency and high-frequency components, respectively, and $n$ is the level of decomposition. A reconstruction filter is then used to reconstruct a filtered signal, $f_1(t)$ from the original signal $f(t)$ and can be expressed as follows.

$$f_1(t) = W * f(t) = A_n(t) + \sum_{j=1}^{n} D_j(t) \quad (8)$$

Where the symbol $W*$ stands for the wavelet transform, a filtering operator in time domain. Then $\|f_1(t)\|_1 < \|f(t)\|_1$ is

satisified, where $\|\cdot\|_1$ is denoted as the one-norm of a time function.

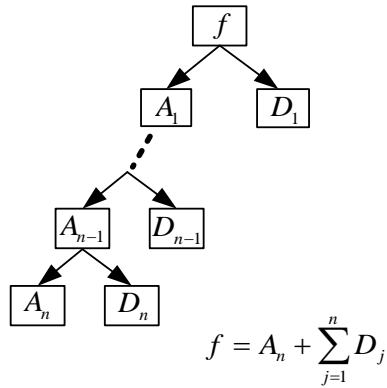

$$f = A_n + \sum_{j=1}^{n} D_j$$

**Figure 3: The wavelet decomposition tree**

*Lemma 1:* K. S. Tzeng and J. S. Chen [8]

Suppose that a signal $f(t)$ can be decomposed as (8) with $\|A_n(t)\|_1 > \gamma \sum_{j=1}^{n} \|D_j(t)\|_1$ where $\gamma > 4$, then there exists a positive real number $\alpha$, such that wavelet operator $(1-\alpha W*)$ is a contraction mapping on $f(t)$, i.e.

$$\|[1-\alpha W*]^k f(t)\|_1 < \rho^k \|f(t)\|_1 \to 0 \quad \text{as} \quad k \to \infty$$
$$\text{for } 0 < \rho < 1 \quad (9)$$

On the basis of practical applications, it is reasonable to suppose that the transfer function $G(s)$ of the controlled system satisfy the properties as shown below [8]:

(A1) The transfer function $G(s)$ has a positive real part for $0 < \omega < \omega_c$ i.e., $\inf_{0<\omega<\omega_c} \text{Re}\, G(j\omega) > 0$, where $\omega_c$ denotes the cut-off frequency of the system.

(A2) The control signal $u_k(t)$ can be decomposed into the learnable part $u_d^l(t)$ and the unlearnable part $u_k^{ul}(t)$, i.e.,

$$u_k(t) = u_d^l(t) + u_k^{ul}(t) \quad (10)$$

(A3) At the k-th iteration, the frequency of the unlearnable part $u_k^{ul}(t)$ is of high-pass type and is bounded by a residual function, $\varepsilon(t)$. After wavelet transform, the unlearnable control input $u_k^{ul}(t)$ satisfy

$$\|W * u_k^{ul}(t)\|_1 \leq \|\varepsilon(t)\|_1, \quad \forall k \quad (11)$$

Where $\|\varepsilon(t)\|_1 \to 0$ as $k \to \infty$

*Remark 1:* The unlearnable part $u_k^{ul}(t)$ of the control signal $u_k^l(t)$ is decomposed into n-level before every iteration, as shown in Figure 3, the unlearnable control input $u_k^{ul}(t)$ will be filtered in the learning loop and will gradually vanish from the learning control signal $u_k^l(t)$ as the iterative learning process goes. Namely, $\|\varepsilon(t)\|_1 \to 0$ when the number of iteration $k \to \infty$.

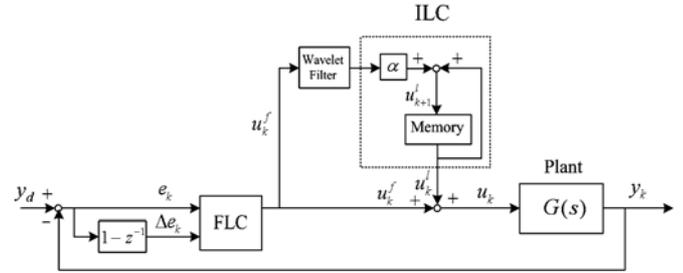

**Figure 4: Structure of the proposed ILC system**

*Lemma 2:* K. S. Tzeng and J. S. Chen [8]

Suppose that an ILC system satisfies assumption (A1)-(A3) and is equipped with the control scheme described as (4)-(6). As $\|\varepsilon(t)\|_1 \to 0$, then there exists a real number $\alpha$, $0 < \alpha < 2$, such that the iterative learning process converges to a positive small number $\tilde{\varepsilon}$, i.e.

$$\lim_{k \to \infty} \|u_k^l(t)\|_1 \to \|u_d^l(t)\|_1 + \tilde{\varepsilon} \quad (12)$$

*Remark 2:* The overall convergent rate will be affected by the choice of learning gain, $\alpha$, and rules of feedback Fuzzy PD controller as well. Namely, these control parameters are determined by the servo bandwidth, $\omega_c$.

Furthermore, while the unlearnable dynamics is vanished, the system dynamics at the $k^{\text{th}}$ iteration can be written as

$$y_k = G(j\omega)u_k \quad (13)$$

where $\omega < \omega_c$.

Let

$$e_k = y_d - y_k \quad (14)$$

Substituting (4)-(6) into (14) to yield

$$e_k = y_d - G(j\omega)(u_k^l + u_k^f) = y_d - G(j\omega)(u_{k-1}^l + \alpha u_{k-1}^f + u_k^f) \quad (15)$$

It implies

$$e_k \left[1 + G(j\omega)K_P + G(j\omega)K_D(1-z^{-1})\right]$$
$$= e_{k-1}\left[1 - (\alpha-1)G(j\omega)K_P - (\alpha-1)G(j\omega)K_D\left(1-z^{-1}\right)\right] \quad (16)$$

Where $\Delta$ is denoted as difference operator, $(1-z^{-1})$.

Therefore,

$$\frac{\|e_k\|}{\|e_{k-1}\|} = \frac{\|1-(\alpha-1)G(j\omega)K_P - (\alpha-1)G(j\omega)K_D(1-z^{-1})\|}{\|1+G(j\omega)K_P + G(j\omega)K_D(1-z^{-1})\|}$$

$$= \frac{\|1+G(j\omega)K_P + G(j\omega)K_D(1-z^{-1}) - \alpha G(j\omega)(K_P + K_D(1-z^{-1}))\|}{\|1+G(j\omega)K_P + G(j\omega)K_D(1-z^{-1})\|} < 1$$

(17)

With $0 < \alpha < 2$, $\omega < \omega_c$, and $\|\cdot\|$ denoted as the two-norm of a time function. Apparently, $\alpha$, $K_P$ and $K_D$ would decide the convergent rate. According to (A1), as $\alpha = 1$, the convergent rate becomes

$$\frac{\|e_k\|}{\|e_{k-1}\|} = \frac{1}{\|1+G(j\omega)K_P + G(j\omega)K_D(1-z^{-1})\|} < 1 \text{ for } \omega < \omega_c$$

(18)

It is noted that, according to Lemma 2, as the feedback control signal $u_k^f$ approaches $\varepsilon$ and the learning control signal $u_k^l$ approaches $u_d^l$ as $k \to \infty$, the stability of the closed-loop control system would be ensured. In addition, if $K_P$ and $K_D$ are the optimal values at the $k^{th}$ iterative, the choice of $\alpha = 1$ will achieve the fastest convergent rate.

*Remark 3:* As $\alpha$ is held constant, i.e., 1, larger $K_P$ and $K_D$ imply smaller steady-state error at every iteration process while the convergent rate remains unchanged as a small $K_P$ and $K_D$ are adopted. Obviously, using a large $K_P$ and $K_D$ can reduce the initial learning error, but it must be limited in order to ensure that the resonant phenomenon of the controlled system would not be excited.

**3.2 The feedback loop-Fuzzy PD controller design [17]**

The Fuzzy PD controller uses two inputs: the tracking error at the $t$ second of $k^{th}$ iteration $e_k(t)$ and its change $\Delta e_k(t)$, where the tracking error can be defined as $e_k(t) = y_d(t) - y_k(t)$, where $y_d$ is the desired output trajectory and $y_k$ is the system output of the $k^{th}$ iteration. So the Fuzzy PD control law can be expressed as follows:

$$u_k^f(t) = F(e_k(t), \Delta e_k(t))$$ (19)

The alternative type of the FLC have two inputs, $e_k(t)$ and $\Delta e_k(t)$, and one output, $u_k^f(t)$. By application of the reasoning algorithm considering $u_k^f(t)$ as an output of the FLC, we arrive at the mapping:

$$u_k^f(t) = f(e_k(t), \Delta e_k(t))$$ (20)

The mapping implemented in the FLC is analogous to the combination of proportional and derivative control, PD controller [18]:

$$u_k^f(t) = K_p e_k(t) + K_D \Delta e_k(t)$$ (21)

Where $K_P$ and $K_D$ are the parameters of PD controller. This control strategy is adopted in the feedback loop to attenuate unlearnable dynamics, alternatively, the Fuzzy PD controller needs to adjust the control effort $u_k^f(t)$ according to $e_k(t)$ and $\Delta e_k(t)$ until they all become zero. Based on above approach, one could formulate rules for all possible cases, as shown in Table 1. Where positive large is denoted as PL, positive small is denoted as PS, zero is denoted as Z, positive small is denoted as NS and negative large is denoted as NL. Note that the body of the table lists the linguistic-numeric consequents of the rules, and the left column and top row of the table contain the linguistic numeric premise terms. Here, with two inputs and five memberships for each of these, there are at most $5^2 = 25$ possible rules.

**Table 1: Rule-base of Fuzzy PD controller for FLC**

| $e_k$ / $\Delta e_k$ | NL | NS | Z | PS | PL |
|---|---|---|---|---|---|
| NL | NL | NL | NL | NS | Z |
| NS | NL | NL | NS | Z | PS |
| Z | NL | NS | Z | PS | PL |
| PS | NS | Z | PS | PL | PL |
| PL | Z | PS | PL | PL | PL |

**3.3 Optimization of FLC using genetic algorithm [19]**

To tune Fuzzy rules, a GA is employed to search for the inference rules that can optimize $K_P$ and $K_D$. A block-diagram using GA in Fuzzy rules tuning is given in Figure 2.

It is known that GA is a search algorithm for problems requiring effective and efficient searching and modeled based on the mechanics of certain natural genetics. GA use the optimizations of reproduction, crossover, and mutation to generate the next generation. From generation to generation, the minimal value of the fitness function output is achieved for each generation. In our case, fitness function, $\Phi_f$, is chosen to be the minimal errors of RMS after $k$ iterations:

$$\Phi_f = \min\left(\sqrt{\frac{1}{N}\sum_{n=0}^{N} e_k^2(nT_s)}\right)$$ (22)

Where time of one iteration, $t$, is converted to $NT_s$; $T_s$ is sampling time, and $N$ is sampling points for every iteration. The population will be improved because fitter offsprings replace parents. The procedure is repeated until either a minimum number of generations are reached or an optimal solution is obtained, whichever is earlier. In this study, we repeat the search procedure until a controller satisfying a pre-specified condition is found. A flow chart of GA algorithm for the inference rules of optimization is shown in Figure 5.

*Remark 4:* It is noted that the GA algorithm would pick the maximum $K_P$ and $K_D$ for the best inference rules. A fitness function based on calculating the minimum RMS for $k$ iterations is used for this purpose. It is also seen from (18) the fastest convergent rate and the minimal RMS can only be obtained under the assumption that the initial RMS to be the same for each iteration, therefore, an augmented PID loop is adopted for this purpose.

*Remark 5:* It is also noted that the augmented PID is activated only when GA algorithm need to select the best rules, once the best rules are determined it would switch to Fuzzy PD controller for iteration process.

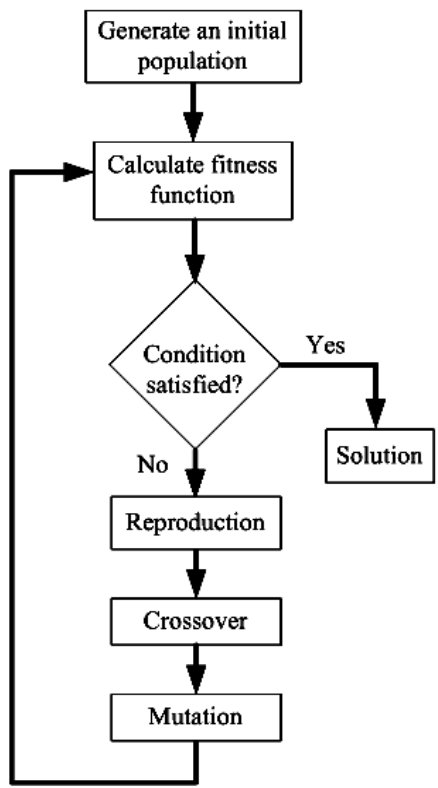

**Figure 5 : A flow chart of GA algorithm for the inference rules searching**

In general, the variables of both input and output membership functions are assumed to be symmetric with respect to the origins. In [19], the scaling factors (SFs) and the deforming coefficients (DCs) are utilized to describe Fuzzy sets for each variable. The effect of the scaling factors is that after all points of the universe of discourse of the Fuzzy PD controller multiplied by SFs, their intervals are changed equally. Then, DCs are utilized to redistribute the location of the membership functions unequally. On the whole, only one side need be calculated because the membership functions are symmetric with respect to the origin as we mentioned above so that we can take symmetrization to create the other side. Therefore, SFs and DCs of the membership functions of input $e_k(t)$, input $\Delta e_k(t)$ and output $u_k^f(t)$ are needed to encode an the Fuzzy PD controller of two inputs and one output in the form of a chromosome for GA. A Fuzzy PD controller of two inputs and one output is encoded in the form of a chromosome with three SFs and three DCs for GA as follows:

$$\begin{bmatrix} S_{I_1} & S_{I_2} & S_{O_1} & D_{I_1} & D_{I_2} & D_{O_1} \end{bmatrix} \quad (23)$$

where using $(S_{I_1}, D_{I_1})$, $(S_{I_2}, D_{I_2})$ and $(S_{O_1}, D_{O_1})$ for Fuzzy PD controller input 1, 2 and output, respectively. By two spaces searched for SFs and DCs from 0.1 to 1 and 0.5 to 0.999, respectively, the scheme can search out the good rules gradually. Through process of 5 generations, the optimal value of the fitness function is $3.9736 \times 10^{-5}$ tantamount to the chromosome

$$\begin{bmatrix} 0.21943 & 0.26153 & 0.9749 & 0.82935 & 0.64842 & 0.64076 \end{bmatrix} \quad (24)$$

Besides, the membership functions for the input and output variables are shown in Fig.6.

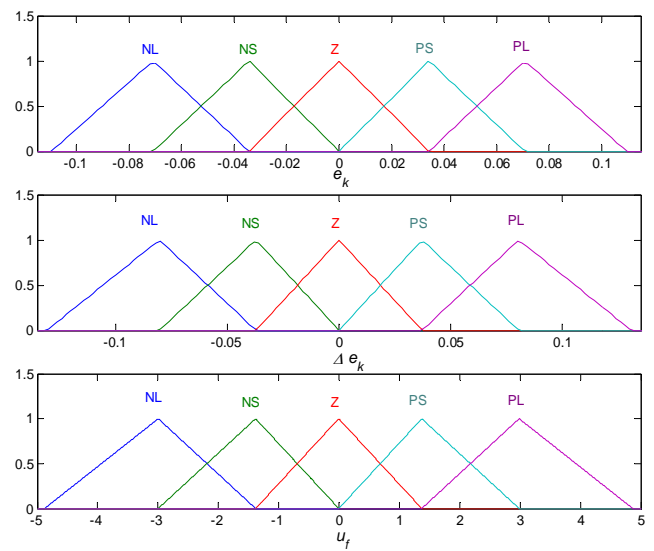

**Figure 6: Input and output membership functions of the best FLC of the 5th generation.**

## 4 Simulation study

Figure 7 shows the learning curves for the fixed Fuzzy rules but different $\alpha$, i.e. $\alpha = 0.1, 0.3, 0.6, 1, 1.4$ and $1.8$. They clearly showed that at $\alpha = 1$, it can achieve the best convergent rate as before. When $\alpha$ is equal to 1, RMS will converge to a small value in the $6^{th}$ iteration as shown in Figure 7.

The simulation results at the 8$^{th}$ iteration as $\alpha = 1$ are shown in Figure 8 and Figure 9. It is noted that the GA algorithm has provided the best rule, therefore, the iteration process would switch to Fuzzy PD controller for the result as shown in Figure 9. The transient response of pneumatic cylinder position is also shown in Figure 8.

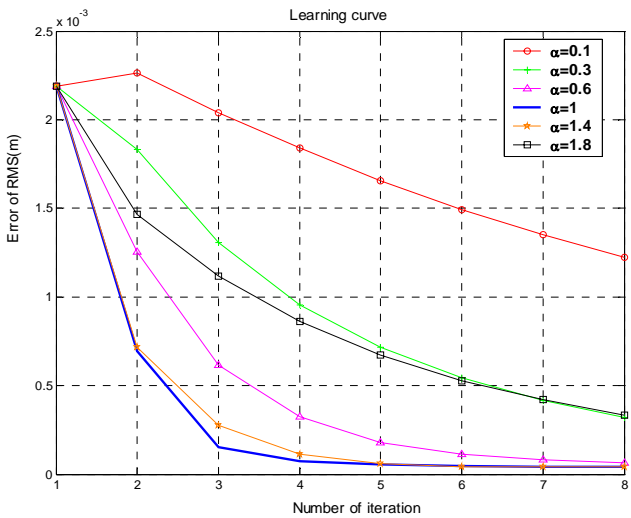

**Figure 7: The learning curve using different learning gains ($\alpha$ =0.1, 0.3, 0.6, 1, 1.4 and 1.8), but fixed Fuzzy rules.**

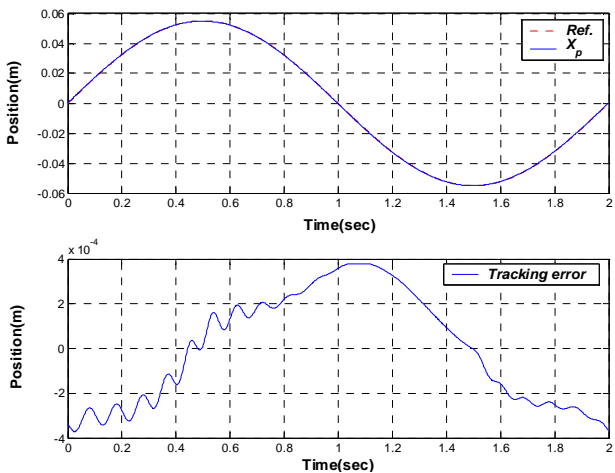

**Figure 8: Transient response of the system under study at the 8$^{th}$ iteration.**

It is noted that the learnable part of $u_k^f(t)$ is extracted by decomposing $u_k^f(t)$ eight times and then reconstructing the low frequency part since learnable dynamics is always at the low frequency region. Since in every iteration, the unlearnable part $u_k^{ul}(t)$, or the non-repetitive part, of the control signal $u_k^l(t)$ has been decomposed into eight-level and all of these signals will be excluded in the reconstructed process of the wavelet transform, the unlearnable control input $u_k^{ul}(t)$ will gradually vanish from the learning control signal $u_k^l(t)$ during the iterative learning process. And, it is seen that the frequency of $u_k^l(t)$ is converged to the reference signal because the reference signal is repetitive and learnable. It can also be seen in Figure 9 that the time response of $u_8^l$ is nearly equivalent to $u_8$. It is also noted that the amplitude of control signal $u_k^l(t)$ is much larger than that of $u_k^f(t)$ because the learnable signal is generated by the reference input of low frequency and the unlearnable signal is caused by unrepeatable dynamic such as noise and it is in high frequency and thus separable. This result can be clearly seen in Figure 9.

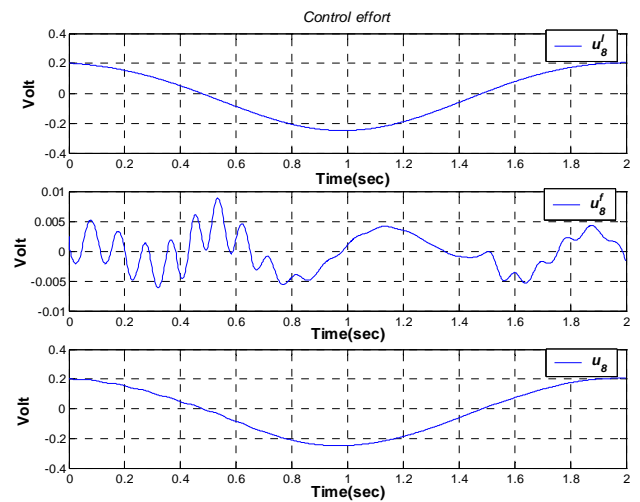

**Figure 9: The Control efforts at the 8$^{th}$ iteration**

## 5 Conclusions

In this article, a WILC and Fuzzy PD feedback control scheme is proposed and its application on the position tracking control of a proportional-valve-controlled pneumatic servo system is investigated The simulation results show that the pneumatic system can follow the reference trajectory efficiently by the proposed controller design. The learnable information of the previous feedback input have been decomposed by wavelet filter and can be used to update the control signal for the current trial and the unlearnable information is attenuated by Fuzzy PD controller such that the tracking errors can be reduced. By WILC with Fuzzy PD feedback control, the system can follow the repetitive reference track after certain learning process. The learning gain and the Fuzzy PD rules are also discussed here as the parameters of the controller. The learning gain $\alpha$ and the Fuzzy PD rules can affect the stability of the controlled system, the RMS convergence speed and the better learning update scheme. The optimization of the Fuzzy PD rules can be found by searching of GA.

**Acknowledgement**

The authors would like to thank for financial support from National Science Council, Taiwan (ROC) for this research under the contract number NSC94- 2218-E007-055.